\documentclass[10pt,twocolumn,letterpaper]{article}

\usepackage{cvpr}
\usepackage{times}
\usepackage{epsfig}
\usepackage{graphicx}
\usepackage{amsmath}
\usepackage{amssymb}
\usepackage{subfigure}
\usepackage{overpic}

\usepackage{enumitem}
\setenumerate[1]{itemsep=0pt,partopsep=0pt,parsep=\parskip,topsep=5pt}
\setitemize[1]{itemsep=0pt,partopsep=0pt,parsep=\parskip,topsep=5pt}
\setdescription{itemsep=0pt,partopsep=0pt,parsep=\parskip,topsep=5pt}

\usepackage[pagebackref=true,breaklinks=true,letterpaper=true,colorlinks,bookmarks=false]{hyperref}

\cvprfinalcopy 

\def\cvprPaperID{} 

\graphicspath{{figures/}}

\begin{document}

\newcommand{\figref}[1]{图\ref{#1}}
\newcommand{\tabref}[1]{表\ref{#1}}
\newcommand{\equref}[1]{式\ref{#1}}
\newcommand{\secref}[1]{第\ref{#1}节}


\title{Landslide Monitoring based on Terrestrial Laser Scanning: A Novel Semi-automatic Workflow}

\author{Yue Pan$^{1}$\\
    $^{1}$ School of Geodesy and Geomatics,Wuhan University \quad \quad\\
}

\maketitle

\begin{abstract}
In this paper, we propose a workflow that uses Terrestrial Laser Scanning(TLS) to semi-automatically monitor landslide and then test it in practice. Firstly, several groups of TLS stations are set on different time to collect the raw point cloud of the object mountain. Next, Hierarchical Merging Based Multi-view (HMMR) registration algorithm is adapted to accomplish single-phase multi-view registration.In order to analyze deformation between multiple periods, Iterative Global Similarity Point (IGSP) algorithm is applied to accomplish multiple-phase registration, which outperforms ICP in experiments. Then the cloth simulation filtering (CSF) algorithm was used together with manual post-processing to remove vegetation on the slope. After that, the mountain slope's digital terrain model (DTM) is generated for each period, and the distance between adjacent DTMs are calculated as the landslide deformation mass. Furthermore, average deformation rate of the landslide surface is calculated and analyzed.To validate the effectiveness of proposed workflow, we uses the TLS data of five periods of the landslide in the Shanhou village of northern Changshan Island from 2013 to 2015. The results indicate that the method can obtain centimeter-level deformation monitoring accuracy which can effectively monitor and analyze long-term landslide morphology and trend as well as position the significant deformation area and determine the type of landslide. In addition, the process can be automated to provide end-to-end TLS based long-term landslide monitoring applications, providing reference for monitoring and early warning of potential landslides.
\end{abstract}

\section{Introduction}\label{sec:Introduction}
Landslides are a major natural hazard associated with surface deformation, mainly occurring in mountainous, hilly, foundation pits, shores and other areas  Landslides would not only cause casualties and property damage in nearby areas, but also cause damage to roads, dams and other infrastructure within its coverage, resulting in traffic stagnation, river breaks and other secondary losses. Therefore, it is of great significance for the monitoring of landslides.

At present, commonly used landslide monitoring methods are listed as following: macroscopic geological observation method, geodetic based method, gravity measurement method, precision liquid static level measurement method, GPS monitoring method, UAV remote sensing monitoring method, interferometric synthetic-aperture radar (InSAR) based method and displacement sensor monitoring method\cite{Pirasteh2016}.The above methods have their own advantages and disadvantages and there is a trade-off between the degree of automation, accuracy, and the number of sampling points. 

Since the 1990s, the novel remote sensing technology laser radar (Lidar) has gradually developed. Among them, terrestrial laser scanning (TLS) is one of the most mature ones \cite{rowlands2003landslide}. TLS is widely used in generating three-dimensional modeling\cite{3DModel}, forest environment monitoring\cite{Forest}, cultural heritage protection\cite{Heritage} and geological disaster monitoring\cite{G2004Terrestrial}.

Recently years,  more and more scholars applied TLS on landslide monitoring \cite{FERNANDEZ2005855} \cite{BALDO2009193}\cite{Prokop2009Assessing}\cite{Oppikofer2009Characterization}. TLS based landslide monitoring apply TLS to provide a time-series of high-resolution point clouds of the topography to generate DTMs and further to understand landslides phenomena \cite{Pirasteh2016}. By comparing the digital elevation information of the point cloud at different periods on the surface of the landslide body, the deformation mass can be calculated\cite{Oppikofer2009Characterization}\cite{Jaboyedoff2012}. Compared with other methods, it has the advantages of high measurement accuracy and the spatial information over the whole landslide area\cite{Landslidedisplacement}. However, due to the influence of terrain, the limited range of scanning angles requires the establishment of multiple stations to scan the entire slope. In addition, point cloud data contains a large amount of noise and includes slope vegetation. Because of the above limitations, the degree of automation of TLS landslide monitoring is limited. 

According to \cite{Prokop2009Assessing} and \cite{Detectcm/mm}, accuracy of TLS-based landslide monitoring mainly depends on: (i) the accuracy and location of the scanner, (ii) the registration quality of the point cloud, and (iii) the filtering degree of slope vegetation, (iv) the generation of a digital terrain model (DTM) and the accuracy of the comparison. \cite{2014TLSTech} suggests that the application of TLS in the field of landslide monitoring can be subdivided into five categories: surface deformation monitoring, volume change estimation, motion velocity analysis, motion mechanism investigation and motion trend analysis. The first three categories belong to the conventional content of landslide deformation monitoring and the latter two categories are the expansion and follow-up of the first three categories.

In view of the limitation on automation of the current TLS based methods and the four main sources of error mentioned in \cite{Prokop2009Assessing}, this paper proposes a new TLS based landslide monitoring workflow that integrates automatic point cloud registration, semi-automatic vegetation filtering and efficient deformation calculation and analysis. The experiment verifies that the workflow can obtain centimeter-level monitoring accuracy and accomplish five type of basic application mentioned in \cite{2014TLSTech}.

\section{Test site}\label{sec:testsite}
\begin{figure}[t]
\begin{center}
\includegraphics[width=1.0\linewidth]{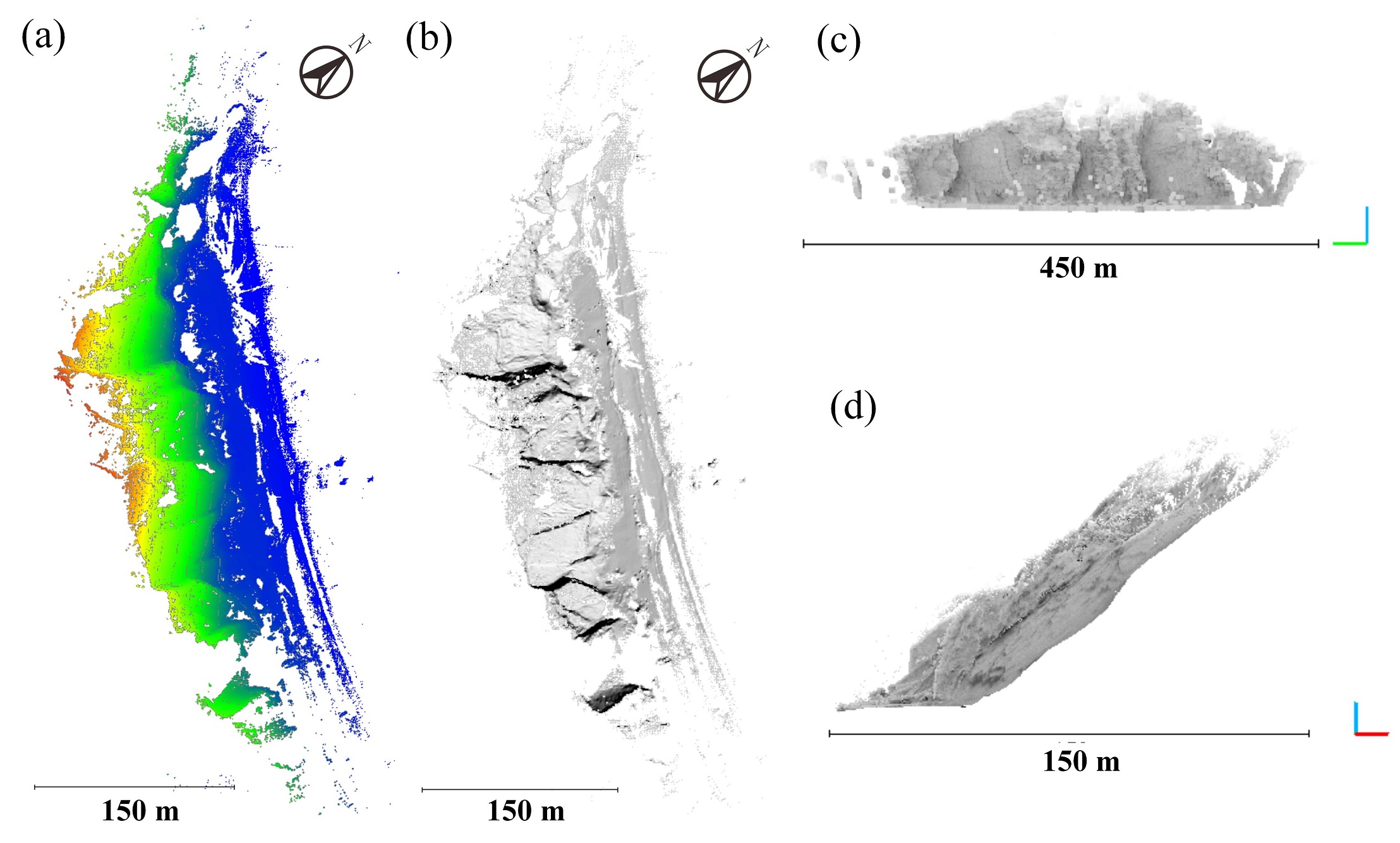}
\end{center}
\caption{Test site profile: (a) top view, elevation coloration (b) top view, shadow coloration (c) main view (d) left view}
\label{fig:4}
\end{figure}

In order to verify the effectiveness of the proposed method, the landslide in Shanhou village of northern Changshan Island is choosed\cite{Shanhou}. As shown in Fig.\ref{fig:13}, The Shanhou Village landslide is located on the southeast side of northern Changshan Island in Shandong Province, P.R.China. The island is a bedrock island, mainly composed of quartzite, which is mixed with slate and phyllite. The joints of the rock mass are perpendicular to the bedding,  resulting in the cliff standing upright with frequent collapse \cite{shanhourock}.  Shanhou village landslide is a rocky landslide with more than 80\% of the slope surface without vegetation cover. Besides, it's the most typical and most harmful landslide in northern Changshan Island. According to \cite{Shanhou}, the quarrying of a working quarry at the foot of the mountain and rainfall are the two important factors causing Shanhou village landslide slip.  As shown in Fig.\ref{fig:4}, the overall width of the landslide is about 450m, the average height is 80m, the total volume is about $5\times 10^5 m^3$, and the slope is nearly $\text{70°}$. From 2013 to 2015, the Leica ScanStation C5 scanner was used to collect the point cloud data close to the mountain. The measurement accuracy was about 6 mm ($m_{TLS}=6mm$) and the average point density is $154pts/m^2$. The observation time and the number of stations are shown in table \ref{tab:1}.

\begin{table}
\caption{TLS setup information}
\begin{center}
\begin{tabular}{|c|c|c|}
\hline
Period&Time&TLS station number\\
\hline
I&2013.03.14&6 \\
II&2013.08.17&4 \\
III&2013.11.06&5 \\
IV&2014.09.13&2 \\
V&2015.01.09&5 \\
\hline
\end{tabular}
\end{center}
\label{tab:1}
\end{table}

\begin{figure}[t]
\begin{center}
\includegraphics[width=1.0\linewidth]{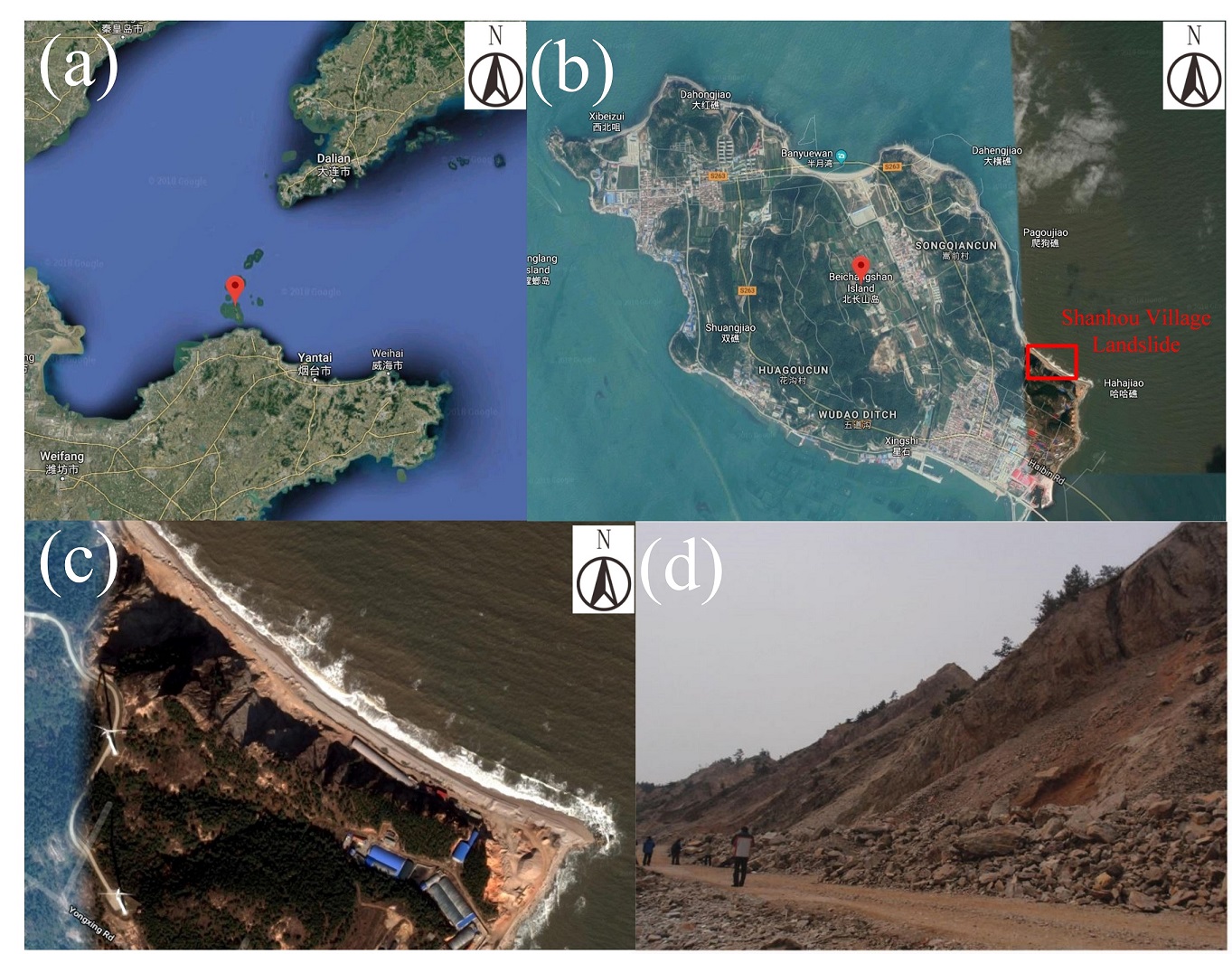}
\end{center}
\caption{Test site: (a)Shandong  Province and Bohai, P.R.China (b)Northern Changshan Island (c)Shanhou Village landslide (d)Test site photo}
\label{fig:14}
\end{figure}

\section{Method}\label{sec:method}
\begin{figure*}[t]
\begin{center}
\includegraphics[width=0.8\linewidth]{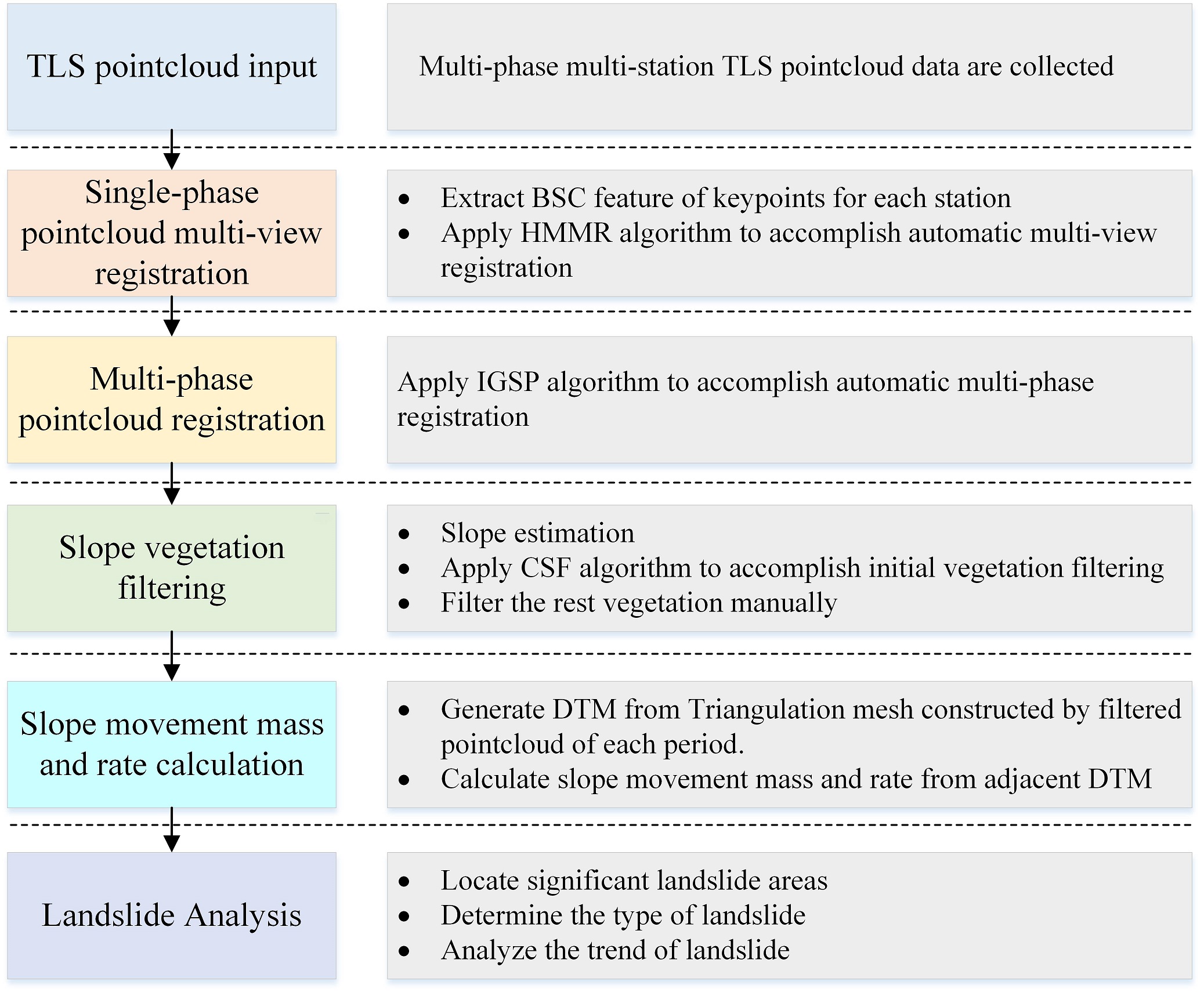}
\end{center}
   \caption{Framework of landslide monitoring based on TLS}
\label{fig:1}
\end{figure*}

The TLS-based landslide monitoring method proposed in this paper follows the process of Fig.\ref{fig:1}. For the input TLS point cloud data, multi-view registration and multi-phase registration are performed first, then the slope vegetation is filtered, and finally the distance between the DTMs generated by each point cloud is calculated to represent the deformation mass. Based on the calculated deformation rate, landslide analysis and further early warning can be done.

\subsection{\textbf{single-phase multi-view point cloud registration}}
Due to the limited range of TLS, in order to obtain complete landslide slope information, multiple stations need to be set up to obtain a three-dimensional point cloud based on the reference frame of each station. The goal of point cloud registration is to unify the point cloud data collected by each station into the same reference coordinate system, and align with each other to obtain complete point cloud data. Therefore, point cloud registration is the basis for all subsequent processing. Generally, the point cloud registration is divided into pairwise registration and multi-view registration. The pairwise registration would register the adjacent two stations with overlapping parts and multi-view registration is accomplished based on pairwise registration. In recent years, scholars worldwide have proposed numerous methods for point cloud registration. There are mainly feature point matching based algorithm \cite{FPFH},  iterative closest points (ICP) based \cite{ICP}\cite{ICP2} algorithm and robust global registration 4PCS based\cite{4PCS}\cite{Super4PCS} algorithm.

\begin{figure}[t]
\begin{center}
\includegraphics[width=1.0\linewidth]{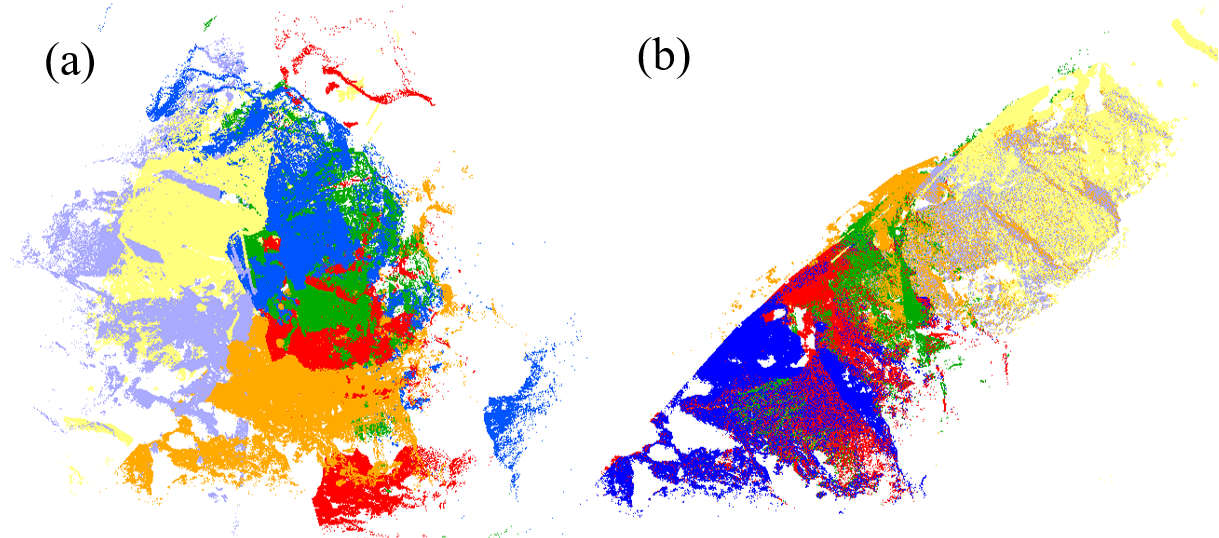}
\end{center}
\caption{Single-phase multi-view point cloud registration based on HMMR: (a) input multi-view point cloud (b) multi-view registration result, different colors represent point cloud collected by different stations.}
\label{fig:2}
\end{figure}

In order to obtain complete information of the mountain slope, multi-view registration needs to be done on multi-station TLS point cloud. In order to obtain an accurate registration result for subsequent landslide deformation calculation,  an algorithm with the highest accuracy is required. Since real-time processing is not required, the algorithm efficiency is not the main consideration. For these reasons, we apply Hierarchical Merging based Multi-view Registration (HMMR) algorithm proposed in \cite{HMMR}.   Firstly,  key points are detected from each station's point cloud. Then the  rotation and translation invariant feature Binary Shape Context (BSC) \cite{BSC} are extracted for each key point.  Next,  HMMR uses feature matching and geometric consistency to filter corresponding key points to complete coarse registration and refine it with ICP to accomplish pairwise registration. Then the local aggregation feature generated by BSCs are used to determine the similarity of any point cloud pair.  Similar point cloud pairs are registered and merged first and the multi-view registration is accomplished hierarchically and iteratively.  Compared with other commonly used registration algorithms, the feature matching and geometric consistency limit improves the accuracy of selected corresponding point pairs. The hierarchical merging method improves the efficiency and accuracy of automatic multi-view registration. According to the original paper \cite{HMMR}, HMMR has an average registration error of about 3 cm ($m_{mreg}=30mm$) on TLS mountain datasets, which satisfies the long-term landslide monitoring needs of this paper. HMMR's result for single-phase multi-view point cloud registration are shown in Fig.\ref{fig:2}.

\subsection{\textbf{Multi-phase point cloud registration }}
After the complete point cloud of slopes have been obtained, in order to compare the difference of point clouds of each period to analyze the deformation and landslide trends, the point clouds of each period also need to be unified to a certain reference coordinate system. Therefore, multi-phase point cloud registration is necessary.  Considering that the cloud data of each period corresponds to the mountain where deformation occurs at different moments, there are local differences in point cloud details. To this end, a robust, fault-tolerant, global registration algorithm is needed. We use the iterative global similarity points (IGSP) algorithm proposed in \cite{IGSP}, which is robust to temporal local difference.  Based on classical ICP,  Euclidean metric is extended to Euclidean and feature hybrid metric, and the global matching strategy based on bipartite graph is used instead of the nearest neighbor of ICP as the corresponding point matching strategy. Through IGSP iterative process, registration is accomplished in a coarse to fine scheme. Although these improvements cause the efficiency of the algorithm to decrease, the robustness of the algorithm is enhanced, which has great advantages for multi-phase point cloud registration which requires global consideration\cite{ICPcompare}.

\begin{figure}[t]
\begin{center}
\includegraphics[width=1.0\linewidth]{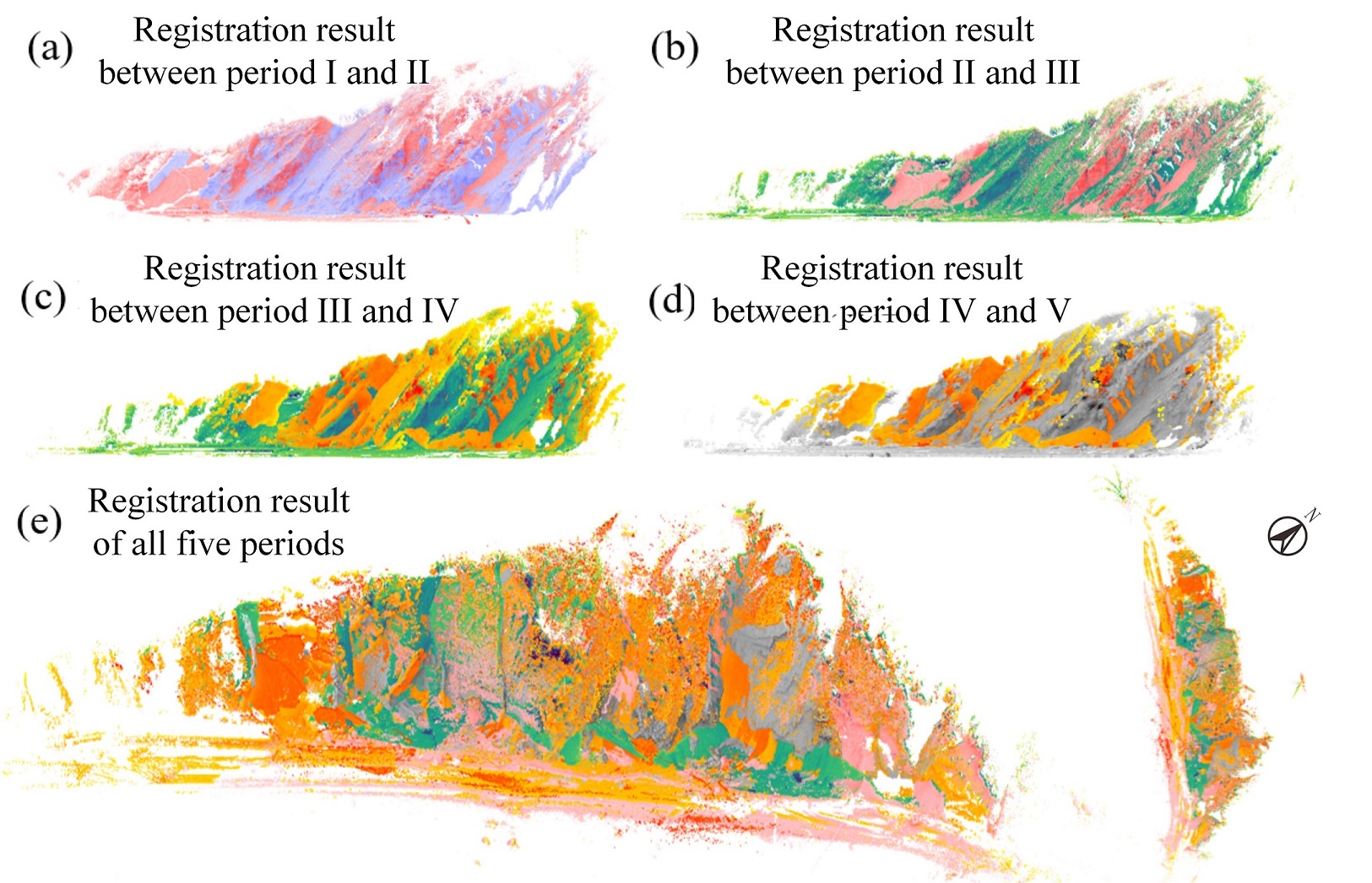}
\end{center}
\caption{Multi-phase point cloud registration results, different colors represent different periods of point clouds}
\label{fig:5}
\end{figure}

In this paper, we apply IGSP, classic ICP\cite{cloudcomparereg} and feature matching with geometric consistency method (FM+GC)\cite{DongReg} to register II, III, IV, V period point clouds with I point cloud respectively. Their rough registration success rate and the corresponding root mean square error (RMSE) is evaluated and compared, as shown in table \ref{tab:2}.  It is shown that IGSP has the highest rough registration success rate and the smallest RMSE, while ICP suffers from
problem of large movements and various point densities\cite{Landslidedisplacement}. Therefore, IGSP is selected and the point cloud of each period is unified to the reference coordinate system of period I. The result is shown in Fig.\ref{fig:5}. According to the original paper \cite{IGSP}, the registration error of IGSP for TLS mountain dataset is about 6 cm ($m_{treg}=60mm$).

\begin{table}
\caption{Comparison of several registration methods for multi-phase point cloud registration }
\begin{center}
\begin{tabular}{|c|c|c|}
\hline
Method& Success rate (\%) & RMSE(m)\\
\hline
IGSP\cite{IGSP}&\textbf{100}&\textbf{0.72}\\
ICP\cite{ICP}&25&3.75\\
FM+GC\cite{DongReg}&75&1.52\\
\hline
\end{tabular}
\end{center}
\label{tab:2}
\end{table}

\subsection{\textbf{Slope vegetation filtering}}
In order to monitor landslide accurately,it is necessary to eliminate non-slope points, such as vegetation, infrastructure, etc. before post-point cloud processing. The filtering algorithm mainly considers the terrain geometric features and the point cloud density. The commonly used morphological methods include least square based method, landslide terrain triangulation based method \cite{2014TLSTech} and the elevation \& slope based raster filtering method \cite{Zhang2003GroundFilter}. However, due to the different geometrical configurations of the slopes, there is no universal mountain vegetation filtering algorithm. For Vegetations whose canopy is high, it is difficult for Lidar pulses to penetrate\cite{su2006influence}. Besides, sparse vegetation would cause confusion between ground and vegetation points. These challenges results in relative low accuracy of vegetation filtering.

\begin{figure}[t]
\begin{center}
\includegraphics[width=1.0\linewidth]{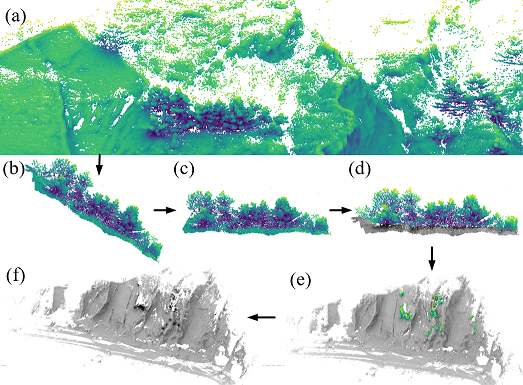}
\end{center}
\caption{Slope vegetation filtration process: (a) original mountain point cloud (b) sub-slope extraction (c) rotate the slope to horizontal plane (d) CSF segmentation of slope points and non-slope points (e) sub-slopes recombination (f) non-slope point (slope vegetation) culling}
\label{fig:6}
\end{figure}

To this end, we propose a mountain vegetation filtering method based on slope segmentation and cloth simulation filtering (CSF) algorithm \cite{Zhang2016CSF}. As shown in Figure \ref{fig:6}, the original mountain point cloud is first rasterized into multiple sub-slopes. For each sub-slope, the average slope inclination can be fitted. Then the slope is rotated into rough horizontal plane. Next, use the CSF-based simple filtering algorithm to segment the slope and non-slope points and finally merge the divided sub-slopes to complete the mountain vegetation filtering. The method effectively solves CSF's problem on hilly areas. However, due to the difficulty in estimating the slope of the geometrically complex region, manual post-processing is still required.

\begin{figure}[t]
\begin{center}
\includegraphics[width=1.0\linewidth]{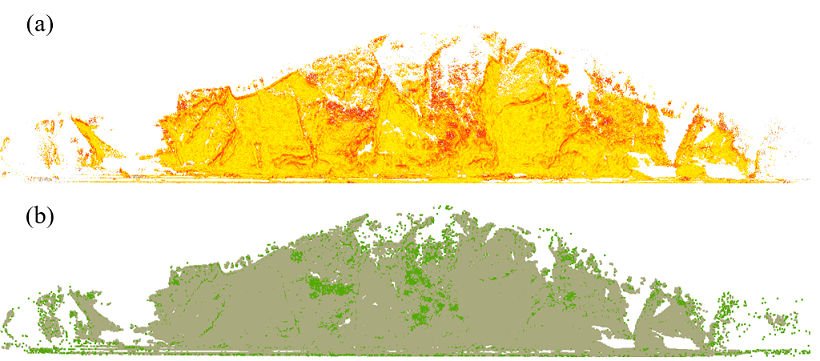}
\end{center}
\caption{Vegetation segmentation algorithm based on visibility gradient: (a) slope visibility gradient vector field map (b) binarization segmentation result based on visibility gradient threshold}
\label{fig:7}
\end{figure}

As a comparison, we also proposes a binary segmentation slope filtering algorithm based on visibility gradient, which works well on the experimental data. The proposed method utilizes the characteristics that the slope vegetation has relatively higher visibility gradient than the slope point under ambient light shielding \cite{Visibility}, and accomplish the segmentation of vegetation and mountain by calculating the visibility gradient and setting a reasonable binarization threshold, as shown in \ref{fig:7}. However, due to the difficulty in adaptive determination of the threshold, the applicability of this method may be limited. In this paper, the CSF-based slope filtering method is finally adopted.

\subsection{\textbf{Slope movement mass and rate calculation}}

For filtered mountain point cloud, in order to analyze the landslide deformation, a common method is to generate DTM by triangulated irregular network (TIN), and then compare it based on DTM\cite{Detectcm/mm}. In this paper, the slope model is generated by using this method.

\begin{figure}[t]
\begin{center}
\includegraphics[width=1.0\linewidth]{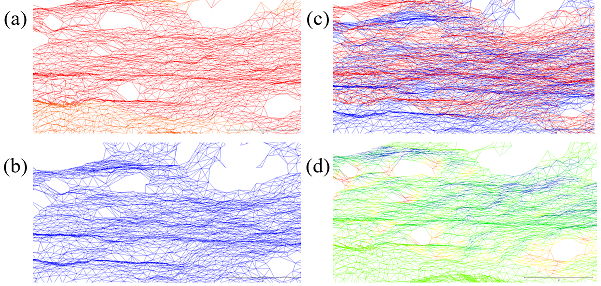}
\end{center}
\caption{Slope model spacing calculation: (a) generate comparison model (b) generate reference model (c) adjacent models (d) distance between adjacent models, shown in gradient color}
\label{fig:8}
\end{figure}

\begin{figure*}[t]
\begin{center}
\includegraphics[width=1.0\linewidth]{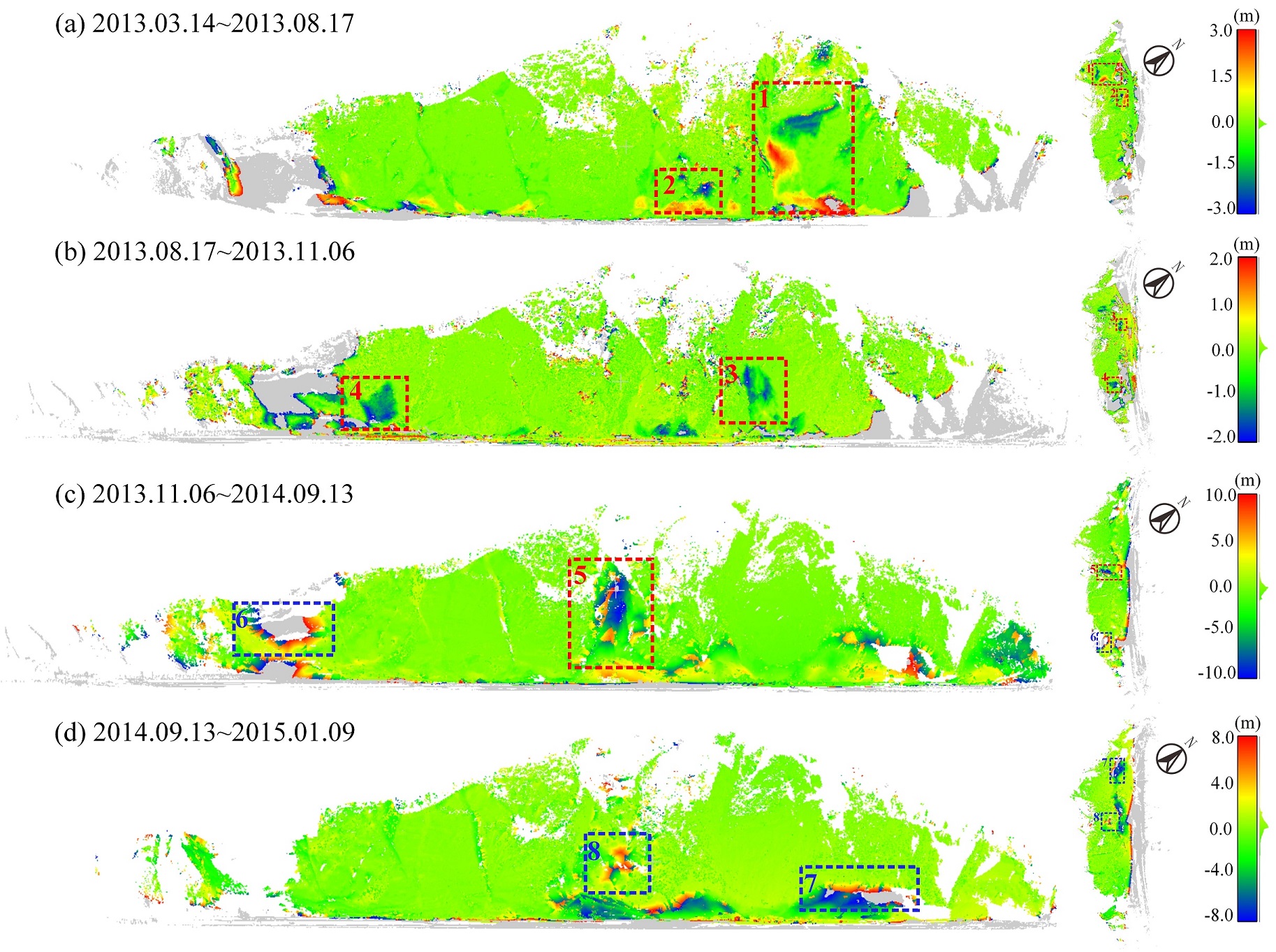}
\end{center}
   \caption{Landslide deformation map of adjacent two periods in front view and top view (deposition area is positive and erosion area is negative). The red frame is the significant landslide area, and the blue frame is the large spacing area caused by incomplete TLS scans.}
\label{fig:9}
\end{figure*}

For two adjacent models, the model spacing can be calculated. In this paper, it is calculated by the model-to-model distance calculation  function of CloudCompare software \cite{CC}, which is achieved by calculating the distance between each vertex of the compared mesh and the reference mesh, as shown in Fig.\ref{fig:8}. It can be used as a good approximation of the actual value of the model spacing given enough high point density. As for the experimental data of five periods, the previous period of the adjacent two models is used as the reference model, and the latter period is used as the comparison model. The approximate spacing of the model is calculated as shown in Fig.\ref{fig:9}, which can be used for representing the deformation of adjacent two periods of landslides. Mean value and standard deviation of displacement are shown in the table \ref{tab:3}.

\begin{figure*}[t]
\begin{center}
\includegraphics[width=1.0\linewidth]{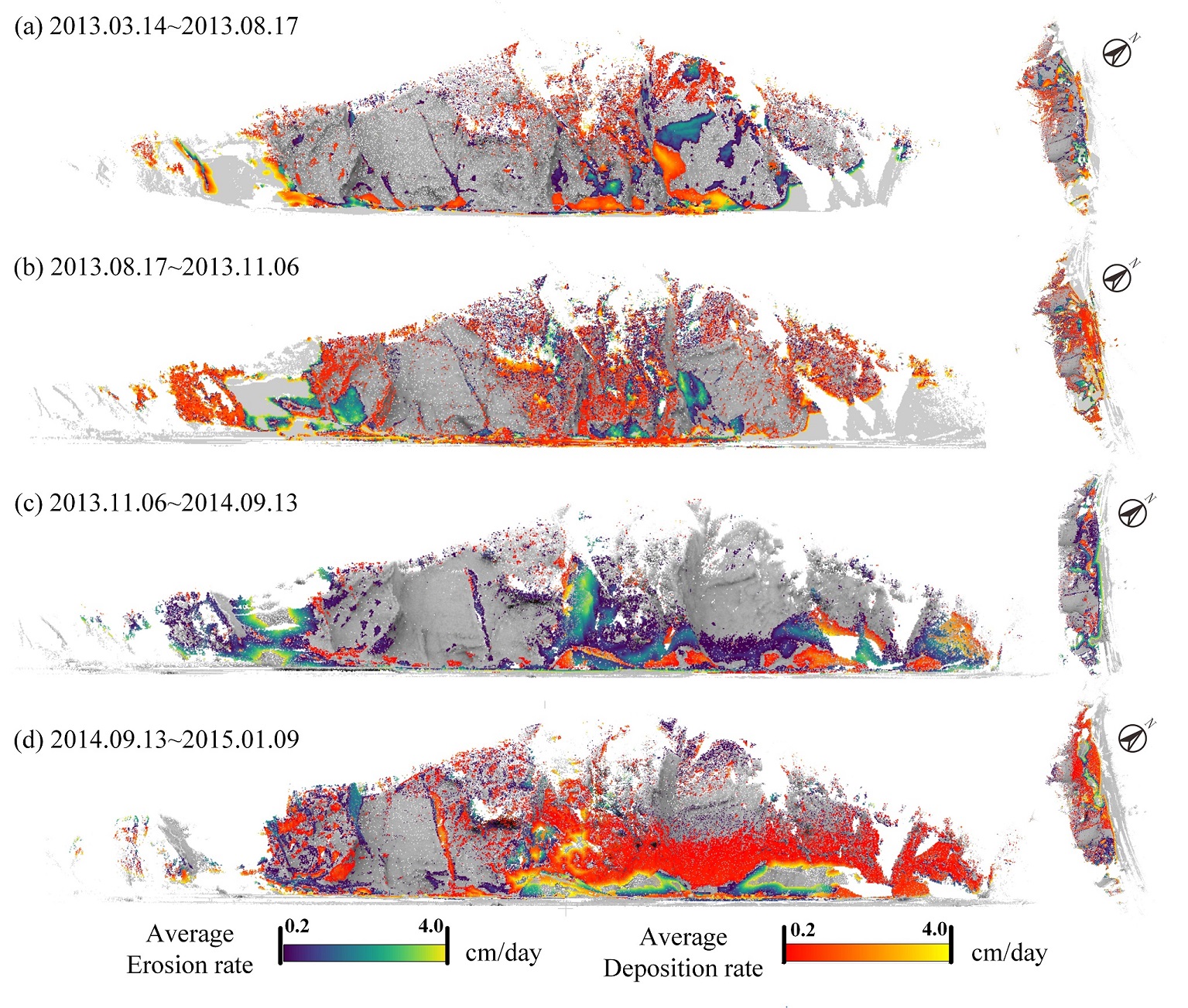}
\end{center}
   \caption{Landslide deformation rate map of adjacent two periods in front view and top view}
\label{fig:10}
\end{figure*}

From the landslide deformation and interval observation time shown in table \ref{tab:3}, the average landslide deformation rate can be calculated. The result is shown in Fig.\ref{fig:10}, highlighting the area where the deformation rate is greater than 2mm/day.

\begin{table*}
\caption{Landslide deformation statistics among five periods}
\begin{center}
\begin{tabular}{|c|c|c|c|}
\hline
period&interval days&mean displacement(cm)&displacement standard deviation(cm)\\
\hline
I,II&156&-0.3&62.7 \\
II,III&81&2.2&41.7 \\
III,IV&311&30.8&158.7 \\
IV,V&118&7.1&217.5 \\
\hline
\end{tabular}
\end{center}
\label{tab:3}
\end{table*}

\section{Analysis and Discussion}\label{sec:4}
\subsection{\textbf{Landslide analysis}}\label{sec:4.1}
In the landslide displacement map shown in Fig.\ref{fig:9}, eight possible significant landslide areas were selected. It has been verified that significant areas 6-8 are mainly caused by the incomplete mountain model due to incomplete scanning. Since point cloud of period IV is only collected by two stations of TLS, a hole is generated in the compared DTM, so the calculated deformation is much larger than the actual value. The other five regions are considered to be actual significant landslide areas.

\begin{figure*}[t]
\begin{center}
\includegraphics[width=1.0\linewidth]{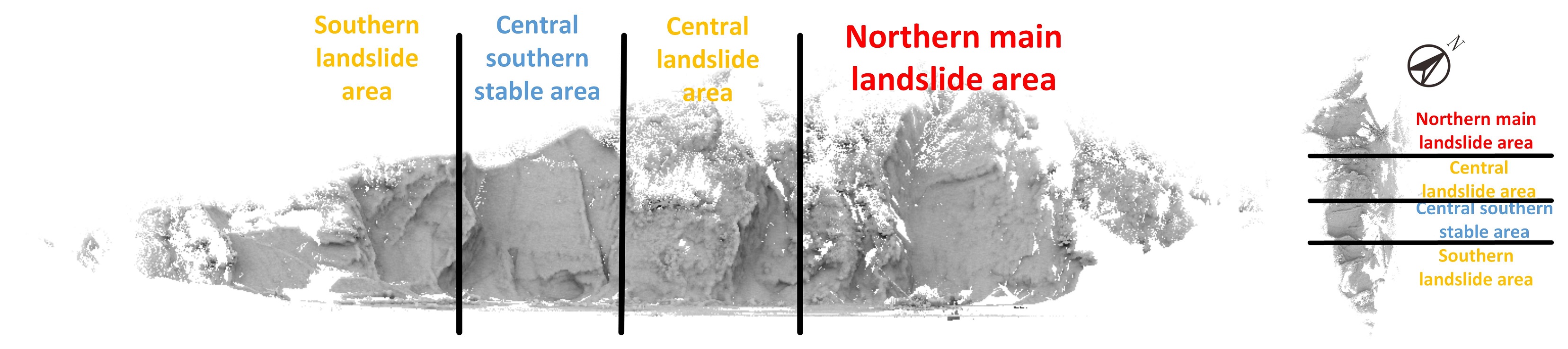}
\end{center}
   \caption{Landslide partition in front and top view}
\label{fig:13}
\end{figure*}

As shown in Fig.\ref{fig:13}, the significant landslide area can be divided into three parts: northern main landslide area (significant area 1-3), central landslide area (significant area 5), and southern landslide area (significant area 4). From period I to period II, the toppling mainly took place in northern main landslide area, and the average deformation rate reached 19 mm/day (see Fig.\ref{fig:10}). From period II to period III, the landslide continued the trend in the northern area. Besides, The landslide occurred in the middle and lower part of southern landslide area, with an average deformation rate of 23 mm/day. From period III to period IV, the landslide mainly occurred in the central landslide area, and the average deformation rate reached 32 mm/day while the landslide trend on both sides was weakened. From period IV to period V, there was no significant landslide in the middle and upper part of the slope, and the average deformation rate was below 10 mm/day. Relatively speaking, the central southern slope is relatively stable between period I and V, whose average deformation rate is below 2 mm/day, called the central southern stable area. From this, the landslide trend can be inferred, the deformation of northern main landslide area and southern landslide area is gradually becoming stable. Central landslide area gradually becomes the main landslide area while central southern stable area keeps being relatively stable.

\subsection{\textbf{Landslide type discussion}}\label{sec:4.2}
Classification of landslides is the basis for studying the formation mechanism and analyzing the trend of future landslides. \cite{Cruden1993Multilingual} proposed a classification criterion based on geotechnical motion characteristics, which divides the landslide into Falls, Topples, Slides, Lateral spreads and Flows. Slides can be further divided into Rotational Slides and Translational Slides \cite{Cruden1996}, which are respectively denoted as FA, TO, S, SP, FL, RS, TS.

\begin{figure}[t]
\begin{center}
\includegraphics[width=1.0\linewidth]{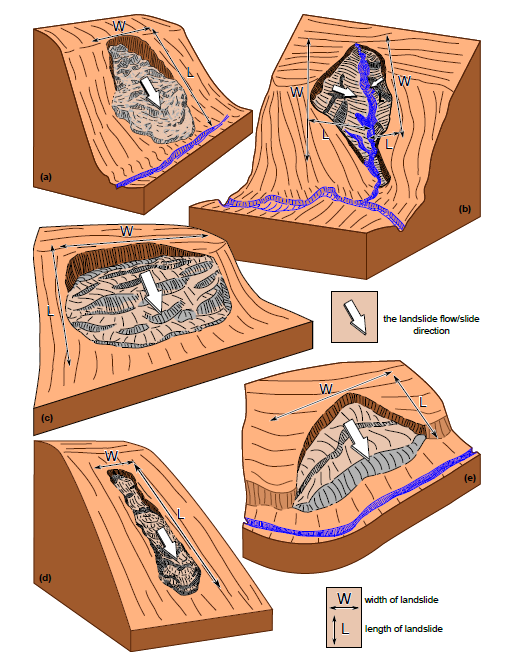}
\end{center}
\caption{Schematic drawings of long and wide cases of landslides\cite{AutomaticlandslidWL}: (a)rotational slide — long type, (b)gully bank slides — long type, (c)translational slide - wide type, (d)flow — long type,
and (e)river bank slide — wide type.}
\label{fig:11}
\end{figure}

Recently, \cite{AutomaticlandslidWL} and \cite {LiJonathan2018} proposed to divide landslides into Very long, Long, Wide and Very wide categories by landslide width and length ratio. In this paper, we denote them as VL, L, W and VW. The length of the landslide (Length: L) is defined as the shortest distance from the tip of the landslide to the top of the toe, and the width (Width: W) is the longest distance between the two wings perpendicular to the landslide moving vector. As shown in formula \ref{eq:1},after calculating landslide shape angle $\theta$, the landslide type can be judged according to formula \ref{eq:2} \cite{LiJonathan2018}. This classification criterion has a certain relationship with the aforementioned classification  of motion characteristics. Flow and most rotational sliding belong to long landslides, while lateral expansion and most translational sliding belong to wide landslides, as shown in fig.\ref{fig:11}.

\begin{equation}
\theta =\arctan \left( \frac{L}{W} \right) 
\label{eq:1}
\end{equation}
\begin{equation}
TYPE=\begin{cases}
	VL,\ \text{67.5°}\leqslant \theta <\text{90° }\\
	L,\ \ \ \text{45°}\leqslant \theta <\text{67.5° }\\
	W,\ \ \ \text{22.5°}\leqslant \theta <\text{45°}\\
	VW,\ \ \text{0°}<\theta <\text{22.5°}\\
\end{cases}
\label{eq:2}
\end{equation}

\begin{figure}[t]
\begin{center}
\includegraphics[width=1.0\linewidth]{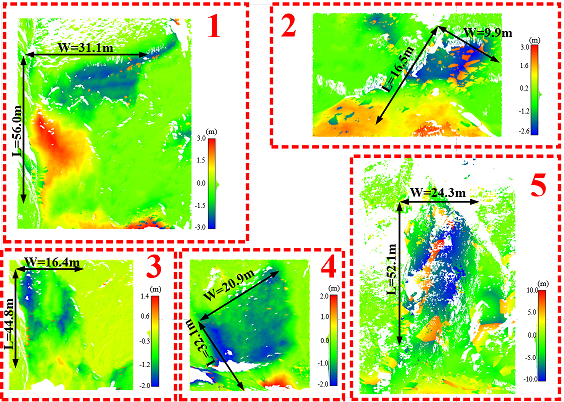}
\end{center}
\caption{Details of five significant landslide areas: landslide deformation displacement, width and length}
\label{fig:12}
\end{figure}

In this paper, for five significant landslide areas identified in \ref{sec:4.1}, the width W and length L are calculated from the deformation area, as shown in \ref{fig:12}. The approximate landslide volume is calculated by model comparison. Then the landslide type is determined by W and L calculation and the geotechnical movement characteristics of the landslide are estimated according to the classification rules proposed in  \cite{Cruden1996}. The results are shown in table \ref{tab:4}. It indicates that the landslide is mainly a long landslide with slides nature. Since there's a working quarry at the foot of the mountain, the mass of the slope keeps decreasing. 

\begin{table}
\caption{Significant landslide area details and classification (type denotation are shown in \ref{sec:4.2})}
\begin{center}
\begin{tabular}{|c|c|c|c|c|}
\hline
Area&W(m)&L(m)&Volume($m^3$)&Type\\
\hline
1&31.1&56.0&648.2&L-RS\\
2&9.9&16.5&213.3& L-RS\\
3&16.4&44.8&445.6& VL-TS\\
4&20.9&32.1&481.5& L-TS\\
5&24.3&52.1&1371.1& L-FL\\
\hline
\end{tabular}
\end{center}
\label{tab:4}
\end{table}
\subsection{\textbf{Error analysis}}
The error $\sigma$ of deformation displacement in this paper can be calculated by the error propagation law, as shown in formula\ref{eq:3}. Where $m_{TLS}$ is the error in TLS measurement, which is approximately 6 mm in this experiment; $m_{mreg}$ is the error in the single-phase multi-view registration, which is approximately 30 mm as previously described; $m_ {treg}$ is the error in multi-phase registration, which is about 60mm; $m_{veg}$ is the error in the residual vegetation of the slope filtering, which is estimated to be 10mm in this experiment; $m_{mesh}$ is the standard deviation of the model-to-model distance calculation which is estimated to be 10 mm in this experiment based on the point density and the upper limit of the triangulation side length. As shown in formula \ref{eq:3}, the measurement accuracy of this method is less than 8cm. For long-term observation whose deformation displacement ranging from 2m to 10m, the relative error is 0.8\% to 4\%, which basically meets the requirement of landslide type classification and trend analysis. The main source of error is the registration during the two periods. Since the registration can be assisted by means of stable points and fixed targets, such accuracy can be improved in the later stage.

\begin{equation}
\begin{split}
\sigma=&\sqrt{2m_{TLS}^{2}+2m_{mreg}^{2}+m_{treg}^{2}+2m_{veg}^{2}+m_{mesh}^{2}}\\
&=\sqrt{2\cdot 6^2+2\cdot 30^2+60^2+2\cdot 10^2+10^2}=76.0mm
\end{split}
\label{eq:3}
\end{equation}

\section{Conclusion and future work}
In this paper, a TLS-based landslide monitoring workflow with high degree of automation and monitoring accuracy is proposed.

TLS data of the mountain near Shanhou Village in northern Changshan Island is used to validate the proposed workflow. The result indicates that it is able to obtain centimeter-level deformation monitoring accuracy and monitor landslides effectively for a long time. At the same time, using these results, the landslide morphology and trend are analyzed, the significant deformation area is located and the landslide type is determined, which indicates that the method can meet the various application needs of landslide monitoring and provide reference for the monitoring of other landslides.

However, the workflow proposed in this paper still has insufficient monitoring accuracy to distinguish centimeter-scale deformation in short term. In order to improve the monitoring accuracy of this method to sub-centimeter or even millimeter level, more registration algorithms assisted by fixed targets and stable points can be tried in the future. In order to get rid of the error of DTM spacing caused by incomplete scanning of the slope, a better arrangement of stations or the assist of Airborne Laser Scanning (ALS) would be tried. As for the landslide morphology, classification, mechanism analysis and susceptibility mapping based on landslide deformation map, the algorithm based on basic GIS operations with the help of landslide inventories \cite{MARTHA2013139} can be exploited to make these tasks more automated.

\paragraph{Acknowledgment} The authors are grateful to Marine Engineering Environment \& Geomatic Center, First Institute of Oceanography, State Oceanic Administration of P.R.China for providing TLS data of Shanhou village in northern Changshan Island, Shandong Province, P.R.China.


{\small
\bibliographystyle{ieee}
\bibliography{Saliency}
}

\end{document}